\documentclass[journal]{IEEEtran}
\usepackage[english]{babel}
\usepackage{graphicx}
\usepackage{psfrag}
\usepackage{amsmath,amssymb}
\usepackage{subfigure}
\renewcommand{\vec}[1]{\hbox{\boldmath{$#1$}}}
\begin{document}

\title{Intelligent location of simultaneously active acoustic emission sources: \\ Part I}

\author{Tadej Kosel and Igor Grabec\\
Faculty of Mechanical Engineering, University of Ljubljana,\\A\v sker\v ceva 6, POB 394, SI-1001 Ljubljana, Slovenia \\e-mail: tadej.kosel@guest.arnes.si; igor.grabec@fs.uni-lj.si%}
\thanks{Manuscript generated: January 31, 2007} }

\maketitle
\begin{abstract}
The intelligent acoustic emission locator is described in Part I, while Part II
discusses blind source separation, time delay estimation and location of two simultaneously active 
continuous acoustic emission sources.
 
The location of acoustic emission on complicated aircraft frame structures is a difficult problem 
of non-destructive testing. This article describes an intelligent acoustic emission source 
locator. The intelligent locator comprises a sensor antenna and a 
general regression neural network, which solves the location problem based on learning 
from examples. Locator performance was tested on different test specimens. 
Tests have shown that the accuracy of location depends on sound velocity and 
attenuation in the specimen, the dimensions of the tested area, and the properties of stored 
data. The location accuracy achieved by the intelligent locator is comparable to that 
obtained by the conventional triangulation method, while the applicability of the 
intelligent locator is more general since analysis of sonic ray paths is avoided.
This is a promising method for non-destructive testing of aircraft frame structures
by the acoustic emission method.  
\end{abstract}

\IEEEpeerreviewmaketitle

\section*{Introduction}
Acoustic emission (AE) concerns non-destructive testing methods and is used to locate and characterize
developing cracks and defects in material.
In non-destructive testing of aviation frame structures, 
acoustic emission is a well accepted method \cite{NTH}. The location 
problem is usually solved by various triangulation techniques based on the 
analysis of ultrasonic ray trajectories \cite{tobias76,chan94,friedlander87}. Solving and programming the related 
equation is rather cumbersome and cannot be simply performed if the 
structure of the tested specimen is geometrically complicated. Acoustic emission testing of aircraft
structures is a challenging and difficult problem. The structures involve bolts, fasteners and plates, all of which
move relative to one another due to differential structural loading during flight. The complex geometry of the 
airframe results in multiple mode conversions of AE source signals, compounding the difficulty of relating the source event
to the detected signal.

In order to avoid difficulties with equation solving and programming of the triangulation 
procedure, several empirical approaches based on learning from examples have already been 
proposed \cite{grabec91}. We developed an intelligent locator capable of learning from examples which we 
therefore called an intelligent locator. The purpose of developing the intelligent locator is to 
replace information obtained from the analysis of sonic ray trajectories by information 
obtained directly from simulated AE events on the specimen under test. In this way, the 
calibration procedure, which has to be performed anyway, could be generalized to the training 
of the intelligent locator.

The development of such an intelligent locator has been described elsewhere \cite{grabec94}. 
In the locator developed a general regression neural network (GRNN) is employed \cite{specht91}, 
which acquires data about the detected AE signals and parameters of their sources during learning. 
The GRNN uses these data in testing when estimating the unknown source position from detected AE signals. 
For this purpose, associative GRNN operation is utilized. The basis of such operation is statistical estimation 
determined by the conditional average \cite{grabecbook}. Consequently, the accuracy of the intelligent locator 
also depends on the learning procedure, and must be examined before testing.

This article describes the results obtained by testing the intelligent locator on experimental 
continuous AE sources. The purpose of this study was to test and examine the advantages of the intelligent locator
compared to a conventional locator. as described in Part I. In Part II an experiment will be explained in which 
an intelligent locator was used to locate two simultaneously active continuous AE sources generated by leakage 
air flow. Location of more than one source at the same time on the test specimen is a new approach in
acoustic emission testing, and is a very promising method for aircraft and airspace structural testing.   

When preparing the experiments, we focused on locating evolving defects in stressed materials and constructions, 
and leakage of vessels. We therefore performed location experiments on four different specimens with three 
different AE sources. The specimens comprised bands, plates, rings, and vessels, while the AE sources were simulated 
by rupture of a pencil lead (pen test), material deformation during tensile test, and leakage air flow through a small 
hole in a sample. 
The positions of AE sources used in testing were well specified. Actual positions were compared with estimated ones, 
and the discrepancy was used to describe the inaccuracy of the locator. In this article, only the experiment with leakage air 
flow through a small hole in a sample is explained. In Part I, location of one continuous AE source is explained.
This Part is intended for better understanding of Part II and comparison of results. 
In Part II, a new approach to the location of two simultaneously active continuous AE sources is explained. 

Below, the article first explains the theoretical background for application of the conditional average to 
the location problem, then describes auxiliary AE signal processing, and finally demonstrates performance of the experimental intelligent locator.

%%%%%%%%%%%%%%%%%%%%%%%%%%%%%%%%%%%%%%%%%%%%%%%%%%%%%%%%%%%%%%%%%%%%%%%%%%%%%%%%%%%%%%%%%%%%%%%%%%
\section*{Theoretical background}
%%%%%%%%%%%%%%%%%%%%%%%%%%%%%%%%%%%%%%%%%%%%%%%%%%%%%%%%%%%%%%%%%%%%%%%%%%%%%%%%%%%%%%%%%%%%%%%%%%
In this section we describe a non-parametric approach to empirical modeling of AE phenomena and solving the 
location problem. This modeling stems from a description of physical laws in terms of probability distributions. 
Since it has been explained in detail elsewhere, we present here just its basic concepts \cite{grabecbook,grabec91}.  
 
The object of empirical modeling is the relationship between variables which are simultaneously measured by a 
set of sensors. In our example the variables are source coordinates and AE signal characteristics.  Let them 
be represented by a vector of $M$ components:  
$\vec{x}=(\xi_1,\ldots,\xi_M)$.  
In the empirical description of an AE phenomenon we repeat the observation $N$ times to create a database 
of prototype vectors $\{\vec{x}_1,\ldots,\vec{x}_N \}$. Instead of formulating a relation between the 
components of $\vec{x}$ we instead treat this vector as a random variable and express the joint probability 
density function $f$ by the estimator   
\begin{align}   
f(\vec{x})=\frac1N\,\sum_{n=1}^{N} \delta(\vec{x}-\vec{x}_n)\,. 
\label{eq16} 
\end{align}  
Here $\delta$ denotes Dirac's delta function. For the purposes of modelling, we must also estimate the probability 
density in the space between the prototype points. This is achieved by expressing the singular delta 
function in Eqs.~\ref{eq16} by a smooth function, such as for example the Gaussian  
\begin{align}   
w_n(\vec{x}-\vec{x}_n,\sigma)=\exp \left [ {{-\Vert \vec{x}-\vec{x}_n \Vert^2}\over 2\,\sigma^2} \right]\,,
\hspace{0.5cm} n=1,\ldots,N\,. 
\label{gauss}
\end{align}  
in which $\sigma$ denotes the smoothing parameter. 
 
The data vectors determine an empirical model of the probability density function. Their acquisition 
corresponds to the learning phase of the empirical modeling. Let us further assume that observation of AE phenomenon 
provides only partial information that is {\em given} by a truncated vector    
\begin{align}   
\vec{g}=(\xi_1,\ldots,\xi_S;\emptyset)\,,  
\label{g}
\end{align}  
in which $\emptyset$ denotes missing components. The problem is to estimate the complementary vector of 
missing or {\em hidden} components:   
\begin{align}  
\vec{h}=(\emptyset;\xi_{S+1},\ldots,\xi_M) ;  
\label{h}
\end{align}  
such that the complete data vector is determined by concatenation    
\begin{align}   
\vec{x}=\vec{g} \oplus \vec{h} = (\xi_1,\ldots,\xi_S,\xi_{S+1},\ldots,\xi_M)\,.
\label{x}
\end{align}  
A statistically optimal solution to this problem is determined by the conditional average estimator, 
which is expressed by a superposition of terms \cite{grabecbook}
\begin{align}  
\hat{\vec{h}}=\sum_{n=1}^{N} B_n(\vec{g})\,\vec{h}_n,\quad {\rm where}\quad \\
B_n(\vec{g})={{w(\vec{g}-\vec{g}_n,\sigma)}\over  
\sum_{k=1}^{N} w(\vec{g}-\vec{g}_k,\sigma)}\,.
\label{nepreg}  
\end{align} 
The basis functions $B_n(\vec{g})$ represent a measure of similarity between the truncated vector 
$\vec{g}$ given by a particular observation and truncated vectors from the database $\vec{g}_n$. 
The higher the value of $B_n(g)$ the higher the contribution of $\vec{h}_n$ to the sum \ref{nepreg} estimating $\hat{\vec{h}}$.
Hence, estimation of the hidden vector $\hat{\vec{h}}$ resembles associative recall, which is characteristic 
of intelligence. The conditional average represents a general non-parametric regression \cite{grabecbook}. 

During the learning phase of operation an intelligent locator of AE sources accepts AE signals and source 
coordinates and stores prototype data vectors, while during application it accepts only AE signals and estimates 
the corresponding source position. Each of these phases can be performed in a separate unit which can be 
interpreted as a layer of a sensory-neural network.  

In order to ensure acceptable properties of the locator, the smoothing parameter $\sigma$ 
must be properly chosen\cite{cherkassky}. The purpose of $\delta$ function smoothing is to estimate the 
probability density function between the prototype data points. A unique method for 
optimal specification of the smoothing parameter is as yet unknown. 
In this case, it is numerically simpler to specify $\sigma$ by the half distance to the 
closest neighbor point:
\begin{align} 
\sigma_n=0.5\,\min_{i} \Vert \vec{g}_{i}-\vec{g}_n \Vert\,, \hspace{0.5cm} \text{for all}\,\, i \ne n\,. 
\end{align}  

%##############################################################################
\subsection*{Signal pre-processing}
\begin{figure}[htb] 
\centering  
\psfrag{X(t)1}[][]{\footnotesize $y_1(t)$}
\psfrag{Y(t)1}[][]{\footnotesize $y_2(t)$}
\psfrag{X(t)}[][]{\footnotesize $y_1(t)$}
\psfrag{Y(t)}[][]{\footnotesize $y_2(t)$}
\psfrag{Rxy}[][]{\footnotesize $R_{y_1 y_2}$}
\psfrag{D}[][]{\footnotesize $\Delta t$}
\psfrag{Cross-}[][]{\footnotesize Cross-}
\psfrag{correlator}[][]{\footnotesize correlator}
\psfrag{Maximum}[][]{\footnotesize Peak}
\psfrag{detector}[][]{\footnotesize detector}
\psfrag{Locator}[][]{\footnotesize Locator}
\psfrag{x}[][]{\footnotesize $\hat{z}$}
\psfrag{Sensor}[][]{\footnotesize Sensor}
\psfrag{Bandpass}[][]{\footnotesize Bandpass}
\psfrag{filter}[][]{\footnotesize filter}
\psfrag{Test specimen}[][]{\footnotesize Test specimen}
\psfrag{#1}[][]{\footnotesize \#1}
\psfrag{#2}[][]{\footnotesize \#2}
\psfrag{H(o)}[][]{\footnotesize $H(f)$}
\psfrag{Signal pre-processing}[][]{\footnotesize Signal pre-processing unit}
\psfrag{Source location}[][]{\footnotesize Source location}
\includegraphics[width=9cm]{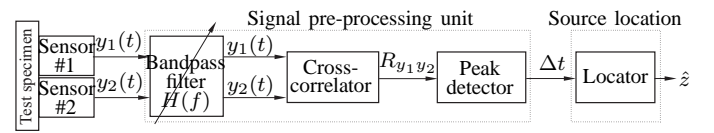}%hosa11.eps
\caption{AE signal processing by the intelligent locator}
\label{potek}
\end{figure}

The intelligent locator comprised a sensor antenna, signal pre-processing unit and source 
locating unit, as shown in Fig.~\ref{potek}. The first unit calculates the time delay $\Delta t$ 
from AE signals $y_1(t)$ and $y_2(t)$, while the second unit estimates the source position 
$\hat{z}$ from the time delay $\Delta t$. AE signals $y_1(t)$ and $y_2(t)$ are detected by 
sensors and filtered using a Butterworth bandpass filter. Without the bandpass filter, 
time delays cannot be easily mapped to source positions on the sample band, and therefore the applicability of this method 
depends on the proper choice of bandpass filter function $H(f)$. We found on dispersive specimens that
information in the continuous AE signal about source position is located in a narrow frequency band.
A wave packet with approximately constant wave velocity along the specimen must be extracted by this filter.  
The filter function $H(f)$ is determined during training procedure of the locator.

Two conventional methods for time delay estimation between two signals are known: threshold function 
and cross-correlation function. 
Estimation of time delay by the threshold function is simple, but only applicable in 
the case of discrete AE. More general, but also more demanding, is time delay estimation 
from the cross-correlation function of AE signals \cite{ziola91}. The cross-correlation function: 
\begin{align} 
R_{y_1y_2}(\tau)=\sum_{t=1}^T y_1(t)\,y_2(t+\tau)\,,
\end{align}  
generally exhibits a peak when parameter $\tau$ corresponds to the time delay $\Delta 
t$ between signals $y_1(t)$ and $y_2(t)$. The time delay is thus determined from the position of the 
peak of the cross-correlation function. One advantage of the application of the cross-correlation function is 
that it does not depend on the discrete or continuous character of AE signals. This method for time delay estimation is
only applicable when one AE source is active at the time of detection. In the event of two or more simultaneously active 
continuous AE sources, a different approach should be used which will be discussed in the Part II.

A filter function is calculated during calibration of the intelligent locator as follows. During calibration, a set of
prototype sources is generated on the test specimen by a pen test at a prepared coordinate net\cite{NTH}. This net 
in most cases has linear sections, where the prototype sources are positioned on a straight line. In this case, we know
that time delays between signals are also linearly dependent. If we have a test specimen with a complicated geometrical structure,
then a pre-calibration process has to be performed in which we have to choose a geometrically simple part of the specimen and carry out a 
pre-calibration procedure on this part such that time delays between signals are linearly dependent. 

\begin{figure}[htb] 
\centering  
\psfrag{X}[][]{\small $l$ [m]}
\psfrag{Y}[][]{\small $\Delta t$ [ms]}
\psfrag{5-15 kHz}[][]{\small 5--15\,kHz}
\includegraphics{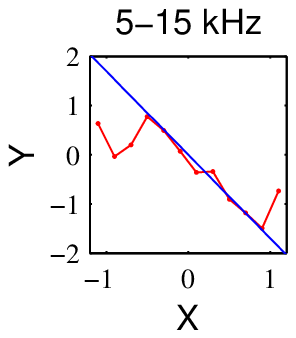}
\hspace{-2mm}
\psfrag{15-25 kHz}[][]{\small 15--25\,kHz}
\includegraphics{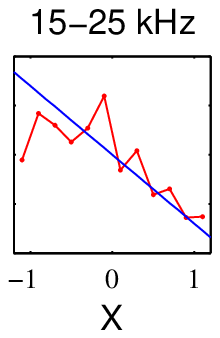}
\hspace{-2mm}
\psfrag{25-35 kHz}[][]{\small 25--35\,kHz}
\includegraphics{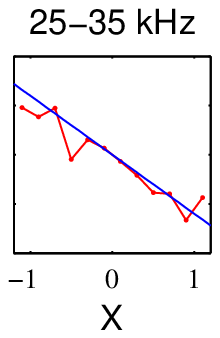}
\hspace{-2mm}
\psfrag{35-45 kHz}[][]{\small \textbf{35--45\,kHz}}
\includegraphics{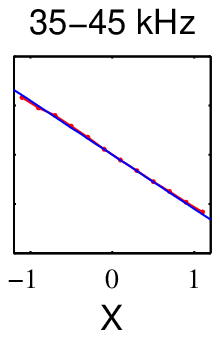}
\hspace{-2mm}
\psfrag{45-55 kHz}[][]{\small 45--55\,kHz}
\includegraphics{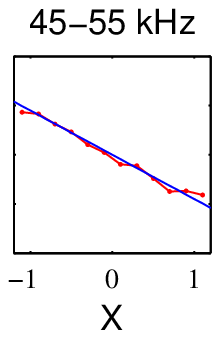}
\hspace{-2mm}
\psfrag{55-65 kHz}[][]{\small 55--65\,kHz}
\includegraphics{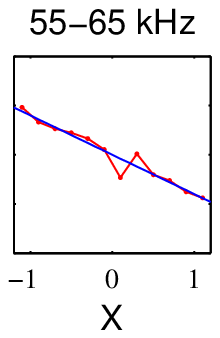}
\caption{Distribution of time delays and their linear approximation along the band specimen. By this procedure an
optimal bandpass filter can be determined.}
\label{dt}    
\end{figure}

\begin{figure}[htb] 
\centering  
\subfigure[]{
\psfrag{x}[][]{\footnotesize $\Delta f$ [kHz] - Frequency band}
\psfrag{y}[b][]{\footnotesize RMSE}
\psfrag{min}[][]{\footnotesize $\Delta f_{\rm opt}$}
\includegraphics{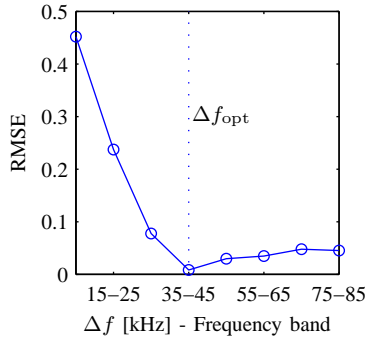}%zdfp.eps
\label{zdfp}
}
\hspace{1cm}
\subfigure[]{
\psfrag{x}[][]{\footnotesize $z$ [mm] - Actual location}
\psfrag{y}[b][]{\footnotesize $\Delta t$ [ms] - Time delay}
\psfrag{-outlier}[][]{\footnotesize -outlier}
\includegraphics{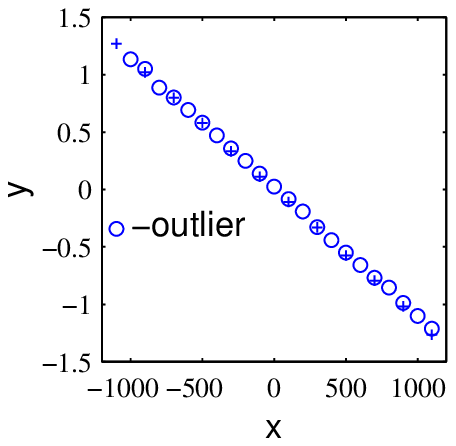}%zveptfp.eps
\label{zveptfp}
}
\caption{Time delays for prototype and test sources by using the bandpass filter of frequency 35-45~kHz. 
{\bf a)} Deviation of prototype source position from a straight line for different filter frequency bandwidth. 
{\bf b)} Time delays of prototype and test sources; Legend:~{\tiny +}~prototype source,~$\circ$~test source}
\end{figure}

For calibration we used AE signals generated by a pen test. We obtained 12 pairs of AE signals from two sensors concatenated 
with known coordinates of sources. The positions of simulated sources were uniformly distributed 
along a straight line on a specimen. In such cases, time delay $\Delta t$ is linearly related to 
source position $z$. This is of advantage for optimal determination of bandpass filter  
because the reference is a straight line.  Calculation of time delays  on the same 
set of prototype AE signals was repeated 70 times. The bandpass filter of 
$\Delta f=10$~kHz was shifted by 1~kHz at each repetition from 5 to 75~kHz. Time delays were 
calculated at each repetition and the distribution obtained was compared with a straight line, as 
shown in Fig.~\ref{dt}. The frequency bandwidth was considered optimal when the root mean square 
error (RMSE) was minimal, as shown in Fig.~\ref{zdfp}. The optimal frequency band for this specimen was 35-45~kHz 
and the velocity of elastic waves was 1.7\,km\,s$^{-1}$. The filter was further used for pre-processing 
samples of prototype as well as test sources. As shown in Fig.~\ref{zveptfp}, the pairs $(z,\Delta t)$, 
estimated from filtered signals, fit a straight line, except one outlier, which results from experimental error.  
 
%###########################################################################
\section*{Experiment}
\begin{figure}[htb] 
\centering  
\psfrag{Senzorji}[][]{\small Sensors}
\psfrag{Operater}[][]{\small Operator}
\psfrag{Analogni}[][]{\small Analog}
\psfrag{signali}[][]{\small Signals}
\psfrag{Kabli}[][]{}
\psfrag{Opazovani sistem}[c][]{\small Specimen}
\psfrag{1}[][]{\small \#1}
\psfrag{2}[][]{\small \#2}
\psfrag{Digitalni}[][]{\small Digital}
\psfrag{osciloskop}[][]{\small oscilloscope}
\psfrag{nastavljanje parametrov}[][]{\small Parameter set}
\psfrag{racunalnik}[][]{\small Computer}
\psfrag{k}[][]{\small Calibration by simulated AE sources}
\includegraphics[width=9cm]{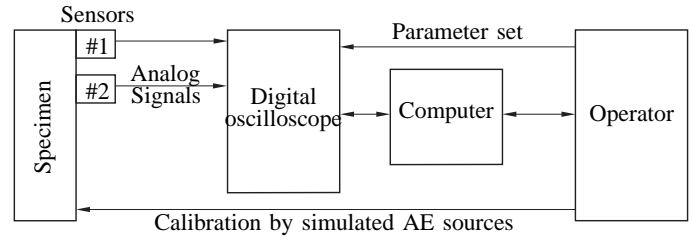}
\caption{Experimental setup of intelligent locator} 
\label{fig3}
\end{figure}

%###########################################################################
The intelligent AE source locator is shown schematically in Fig.~\ref{fig3}. 
It includes an automatic data-acquisition system controlled by computer and 
  a network of AE sensors. 

The AE sensors are piezoelectric transducers (pinducers). The diameter of the transducer active area is 1.3\,mm,  
And so it can be considered a point-like sensor. The signals from sensors are fed to a digital
oscilloscope where they are digitized and transferred to a PC. Operation of the intelligent locator 
is determined by software in the PC that controls data acquisition and estimates the position of unknown AE sources. 

The locator operates in two different modes:
\begin{enumerate} 
\item In learning or calibration mode, a set of $N$ pen tests is performed in which complete information about the 
AE phenomenon is acquired. The operator must prepare an orientation net the shape of which depends on the shape of the
test specimen. The recommended shape is an equidistant net, since such position of prototype sources yield a minimum error 
of the locator. 
>From source coordinates and time delays between pre-processed AE signals, 
the prototype vectors are created and stored in the memory of the neural network as a data base. 

\item In application mode, only time delays between AE signals are provided. There are then associated
in the neural network with the estimated source coordinates.    
\end{enumerate}

In the case of discrete AE, the time delay can visually 
be estimated from a marked jump in the burst of the AE signal, or can be instrumentally determined using a 
threshold function. Hence, in the case of continuous AE, time delays cannot be simply estimated, although a 
cross-correlation function has already been used for this purpose. In our approach, we therefore applied a 
cross-correlation function.  The purpose of this experiment was to determine the accuracy of location of continuous 
AE sources on a one-dimensional specimen. 

Two experiments on aluminum band specimen are explained in this article.
We tested the locator on an aluminum band specimen of dimensions $4000\times 40\times 5$\,mm$^3$. Reflection of AE signals at the ends 
of the band specimen was reduced by sharpening the ends. For testing we selected a test area in the middle of the 
band specimen where 23 holes were prepared. The distance between holes was 100\,mm and the diameter of holes was 2\,mm.  
Two AE sensors were mounted 100\,mm away from the terminal holes. 
For the purpose of locator training, we generated 12 prototype sources separated by 200\,mm,  while all 23 
holes were applied for locator testing. In this experiment, we calibrate the locator by pen test
and examine it by continuous AE generated by air flow. The air flow was produced by expansion of compressed air through
nozzle of 1\,mm diameter. The nozzle was mounted 1\,mm above the band specimen surface.   

Two experiments were performed. In the first experiment, only one continuous AE source was active on the 
band specimen, while in the second experiment two continuous AE sources were active simultaneously on the band specimen. 
Successive simultaneous location of two sources is explained in Part II.  \

Signals were processed as shown in Fig.~\ref{potek}. The first step in processing was calculation 
of cross-correlation function of AE signals. The corresponding signal was sent through a bandpass Butterworth 
filter of bandpass from 35 to 45 kHz. Determination of this filter is explained earlier in this article. 

%RESULTS
%%%%%%%%%%%%%%%%%%%%%%%%%%%%%%%%%%%%%%%%%%%%%%%%%%%%%%%%%%%%%%%%%%%%%%%%%%%%%%%
\section*{Results}
\begin{figure}[htb] 
\centering  
\subfigure[]{
\psfrag{x}[][]{\footnotesize  $x$ [mm] - Actual location}
\psfrag{y}[b][]{\footnotesize $\hat{x}$ [mm] - Estimated loc.}
\psfrag{-outlier}[][]{\footnotesize -outlier}
\includegraphics{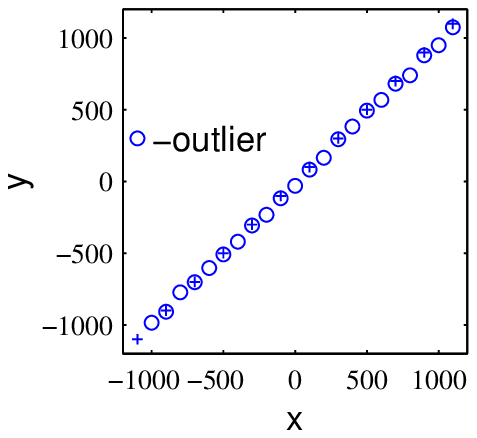}%zverez.eps
\label{zverez}
}
\hspace{1cm}
\subfigure[]{
\psfrag{x}[][]{\footnotesize $x$ [mm] - Actual location}
\psfrag{y}[b][]{\footnotesize $\varepsilon$ [mm] - Absolute error}
\psfrag{ep}[][]{\footnotesize $\varepsilon_a$}
\psfrag{-outlier}[][]{\footnotesize -outlier}
\includegraphics{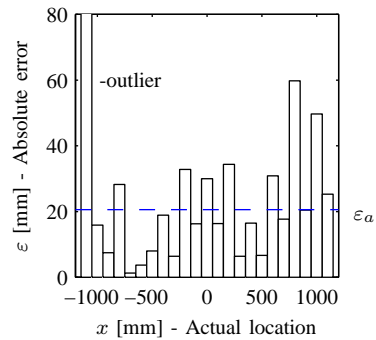}%zveerror.eps
\label{zveerror}
}
\caption{Result of continuous AE source location on the band. {\bf a)} Estimated versus actual location of test sources; Legend: {\tiny +}~prototype source, {\tiny $\circ$}~test source. {\bf b) } Absolute location error; $\varepsilon_a$ - average error.} 
\label{averezerror}
\end{figure}

The results of locator testing are shown in Fig.~\ref{zverez}. The absolute location error for each test source is shown 
in Fig.~\ref{zveerror}. 
%data
Location error in the experiment ranges from 1.3\,mm to 60\,mm with average value $\varepsilon_a=20$\,mm 
(ignoring the outlier). If we describe the error with respect to the distance between sensors (2.4\,m), the relative 
value is less than 1\%. Increasing the number of prototype sources can reduce the error. 
%Topic sentece
Despite the complexity of continuous AE signals, the location problem was solved satisfactorily with respect to 
The accuracy required in non-destructive testing. Results also 
show that a standard calibration procedure with discrete AE signals generated by pen test
can be used for locator training.

%##############################################################################
\section*{Discussion and Conclusion}
%##############################################################################

Estimation of source coordinates by the conditional average is subject to systematic error caused by smoothing of the 
delta function \cite{grabec91}. This error can be reduced by increasing the number of prototype sources.
Since it is not always possible to increase the number of prototype sources due to the
complexity of experiments, a compromise must be found by trial and error. 

Experimental error is acceptable, so we decided to make additional tests, as will be
discussed in Part II.

This study shows that a conventional AE locator operating on the triangulation method 
can be successfully replaced by an intelligent locator that learns from examples.
The results show that the intelligent locator can locate sources with acceptable accuracy in cases of: (1) discrete AE 
on band and plate, (2) continuous AE on band, (3) discrete AE on 
plate with hole (ring), (4) discrete AE generated by specimen rupture during the tensile test, 
and (5) discrete AE on pressure vessel. Is has been also shown that the locator can perform zonal locating\cite{kosel98b}.

Comparing mean errors of all experiments and the distances between prototype sources, 
we find that the average error is always less than 30\% of 
the distance between prototype sources, while the maximal error is always less than 50\% of the distance 
between prototype sources. The accuracy of the locator can be controlled by the number of prototype 
sources excited during training. The experimental error of the locator is a consequence of wave dispersion on a specimen
that operates as a waveguide, 
reflections from boundaries, and attenuation. 
We found for dispersive waves that an optimal wave packet must be found which has approximately constant velocity 
along the test specimen.
Estimation of time delay  between AE signals
by the cross-correlation function is only applicable for one active AE source. If there are several simultaneously active AE sources,
then blind source separation should be used, as will be shown in Part II.

\end{document}